\documentclass[10pt,aps,prl,twocolumn,showpacs,superscriptaddress]{revtex4-1}  

\usepackage[modulo,switch]{lineno}
\modulolinenumbers[1]

\usepackage[utf8]{inputenc}
\usepackage{graphicx}
\usepackage{amsmath}
\usepackage{amssymb}
\usepackage{bm}        
\usepackage{txfonts}
\usepackage{multirow}
\usepackage{mathptmx}  
\usepackage{siunitx}

\usepackage{xargs}                      

\usepackage[%
  colorlinks=true,
  urlcolor=blue,
  linkcolor=blue,
  citecolor=blue
]{hyperref}

\begin{document}

\title{Increased performance in DDM analysis by calculating structure functions through Fourier transform in time}

\author{M.~Norouzisadeh}
\affiliation{Universite de Pau et des Pays de l'Adour, E2S UPPA, CNRS, TOTAL, LFCR UMR5150, Anglet, France}
\altaffiliation{ Universit\'e Orl\'eans, ISTO, CNRS, 1A rue de la F\'erollerie, 45071 Orl\'eans, France}
\author{G.~Cerchiari}
\affiliation{Institut f\"ur Experimentalphysik, Universit\"at Innsbruck, Technikerstrasse~25, 6020~Innsbruck, Austria}
\altaffiliation{Corresponding author: giovanni.cerchiari@uibk.ac.at}
\author{F.~Croccolo}
\affiliation{Universite de Pau et des Pays de l'Adour, E2S UPPA, CNRS, TOTAL, LFCR UMR5150, Anglet, France}

\date{\today}

\begin{abstract}
Differential Dynamic Microscopy (DDM) is the combination of optical microscopy to statistical analysis to obtain information about the dynamical behaviour of a variety of samples spanning from soft matter physics to biology. In DDM, the dynamical evolution of the samples is investigated separately at different length scales and extracted from a set of images recorded at different times. A specific result of interest is the structure function that can be computed via spatial Fourier transforms and differences of signals. In this work, we present an algorithm to efficiently process a set of images according to the DDM analysis scheme. We bench-marked the new approach against the state-of-the-art algorithm reported in previous work. The new implementation computes the DDM analysis faster, thanks to an additional Fourier transform in time instead of performing differences of signals. This allows obtaining very fast analysis also in CPU based machine. In order to test the new code, we performed the DDM analysis over sets of more than 1000 images with and without the help of GPU hardware acceleration. As an example, for images of $512 \times 512$ pixels, the new algorithm is 10 times faster than the previous GPU code. Without GPU hardware acceleration and for the same set of images, we found that the new algorithm is 300 faster than the old one both running only on the CPU.
\end{abstract}

\maketitle

\section{Introduction}
\label{intro}
In the latest decade, Differential Dynamic Microscopy (DDM) has gained popularity in the field of soft matter physics due to its robustness and easy implementations in all the laboratories already equipped with a microscope~\cite{Cerbino2008,Germain2016,Cerbino2017}. The technique allows investigating the dynamics of rather different samples ranging from colloidal particles~\cite{Cerbino2008} to bacteria \cite{Wilson2011}, but also from biological cells~\cite{Sentjabrskaja2016} to density fluctuations in and outside thermal equilibrium~\cite{Giavazzi2016,Croccolo2007} and many others as witnessed in several review articles~\cite{Cerbino2017,Brader2010,Poon2012}.
Other techniques make use of similar image analysis, like the dynamic shadowgraph that some of us currently use to investigate non-equilibrium fluctuations in complex fluids out of equilibrium and to extract information about transport properties of the mixtures~\cite{Croccolo2007,Croccolo2012,Croccolo2019a,Croccolo2019b}. 

The implementation of DDM requires, as stated, a microscope to acquire series of images by transmitted light~\cite{Cerbino2008}, fluorescence-based~\cite{Peter2012}, dark-field~\cite{Bayles2016} or any other visualisation scheme. The series of images need then to be treated by custom-made software to compute the so-called structure function, as defined by Schultz-Dubois and Rehberg~\cite{Schulz-DuBois1981} and implemented to Schlieren~\cite{Croccolo2006a} and shadowgraph~\cite{Croccolo2006a} imaging and to optical microscopy~\cite{Cerbino2008,Giavazzi2009}.
The different available software share common features and in particular calculate the difference of images and then Fourier transform that signal in space over the 2D~\cite{Germain2016}, or in other cases they first compute 2D-FFT and after compute differences of the image FFTs~\cite{Cerchiari2012}.

Since the amount of acquired images and their 'weight' have considerably increased in the latest two decades, the computational time needed to compute the structure function should have also increased. Of course, also the computational capabilities of modern computers have largely increased, but a major breakthrough has been achieved when researchers have started to implement the computation of the structure function on graphic processor units (GPU)~\cite{Cerchiari2012}.
The implementation of the cited computational task on GPU allowed a decrease of the computational time by a factor 10--30; thus, reducing the data analysis time from several hours to a few tens of minutes.

In the present article, we present a further optimisation step consisting in performing the calculation of the structure function by Fourier transform in time rather than calculating differences of spatial FFTs. This implementation has also been tested on the Central Processing Unit (CPU) in order to test the GPU acceleration in different experimental conditions.

The source code of the program developed for the tests reported in this work, which executes the algorithm both for CPU and GPU, is released under the GNU General Public License v.3~\cite{gplv3} and is freely available for download at~\cite{diffmicro2020}.

\section{The algorithm}
\label{sec:1}
In DDM experiments, the light modulations generated by the sample are acquired by a camera in the form of a set of images $I_n$ at regular time intervals. To measure the dynamical and spatial properties of the specimen, the recorded $N$ images are processed to calculate the structure function $d$. The structure function is defined as follows~\cite{Croccolo2006a}:
\begin{equation}
\label{eq:tobedone}
    d\left(m\right)=\frac{1}{N-m}\sum_{n=m}^{N-1}\lvert F_{xy} \left(I_{n-m}-I_{m}\right)\rvert^2 \; ,
\end{equation}
where the indices $n$ and $m$ run from $0$ to $N-1$ and $F_{xy}$ indicates the bi-dimensional fast Fourier transform (FFT) of the images. The absolute value operation $\lvert \ldots \rvert$ is intended for every wave vector component of the FFT.

In previously described implementations~\cite{Cerchiari2012}, computing eq.~\ref{eq:tobedone} was approached via a two-step algorithm. First, all FFTs of the images $\tilde{I}_n = F_{xy}I_n$ were calculated and stored in the local memory. Second, each structure-function for a specific time delay $m$ was evaluated via an in-place average scheme updating the matrices $d\left(m\right)$ while calculating the differences of image FFTs $\left(\tilde{I}_{n-m}-\tilde{I}_n\right)$. This algorithm exploited the linearity of the FFT operation and the available hardware memory to reduce the computational load and achieve a faster execution time than by applying eq.~\ref{eq:tobedone} directly. The speedup was possible because the FFTs of the images could be re-used several times for different $m$ values instead of calculating a FFT for every difference $\left(\tilde{I}_{n-m}-\tilde{I}_n\right)$. After the FFTs were calculated and saved, the wave vector components of the FFTs remained independent of one another. Each wave vector was elaborated independently by a parallel algorithm designed for graphic processing units (GPU) to efficiently perform the operation of difference, squared modulus and average.

In this work, we present a modified algorithm for computing the operations on the wave vectors, \textit{i.e.} the operations after the image bi-dimensional FFTs are evaluated and stored in the local memory. The new approach optimizes the calculations by computing an additional Fourier transform of the data in time. To describe the new algorithm, we will refer to the vector obtained by concatenating the complex amplitudes of subsequent FFTs at a specific wave vector in the bi-dimensional Fourier space of spatial frequencies with the name \textit{time sequence}. The computation can be described for a single time sequence and iterated over all the wave vectors to obtain the final result. We will show that the new algorithm is suited for both GPU and CPU hardware platforms.

We expand the square modulus operation of eq.~\ref{eq:tobedone} to obtain:

\begin{align}
    \label{eq:newalg}
    d\left(m\right)&=\frac{1}{N-m}\sum_{n=m}^{N-1}\left(\lvert \tilde{I}_{n-m}\rvert^2+\lvert \tilde{I}_{m}\rvert^2-2\textrm{Re}\left( \tilde{I^*}_{n-m}\tilde{I}_{m}\right)\right) \; ,
\end{align}

where the apex ``$*$'' indicates complex conjugation. The first two terms in the sum are averages of the modulus-squared elements of each time sequence. The first term containing $\lvert \tilde{I}_{n-m}\rvert^2$ is the average of the first $N-m$ squared elements of the sequence. Likewise, the second term containing $\lvert \tilde{I}_{m}\rvert^2$ is the average of the last $N-m$ components. Both averages have a computational complexity of $O\left(N\right)$. The last term identified by the product $\tilde{I^*}_{n-m}\tilde{I}_{m}$ is the real part of the auto-correlation of the time sequences. As widely known in signal processing, the auto-correlation can be evaluated via the power spectrum by using the FFT operation over the time sequence. The advantage of computing the auto-correlation via Fourier transform in time is the speed-up of the FFT algorithm. Given $N$ images, with $N$ being a power of two, the FFT transforms $O\left(N\times N\right)$ operations (the summation over $n$ for all $m$), into $O\left(2 \times N\times\log_2\left(N\right)\right)$ plus $O\left(N\right)$ (the direct and inverse FFTs plus the elementwise square modulus). Thus, it is possible to reduce the computational complexity of the algorithm from $O\left(N^2\right)$ to $O\left(N\times\log_2\left(N\right)\right)$ by calculating the auto-correlation with the help of Fourier analysis of the data over time. More details about our Fourier analysis approach over time are reported in Appendix B.

To compare the new algorithm with the one described in Ref.~\cite{Cerchiari2012}, we prepared a program that implements both algorithms on CPU and GPU hardware for a total of four execution modes. To distinguish the two algorithms we will refer to the method reported in Ref.~\cite{Cerchiari2012} as WITHOUT\_FT and the technique discussed in this article as WITH\_FT. Both methods calculate the final result in two steps. The first step is common and consists in calculating and storing the FFTs of the images in the available free memory: RAM for the CPU versions and global G-RAM for the GPU implementations. In the second step, the wave vectors are analyzed independently according to the different schemes. If the wave vector data exceeds the capacity of the available memory, both algorithms split the job into several groups at the price of recalculating the image FFTs several times (see Appendix C for more details). The program is written in C++11 and CUDA~v.10.2 with graphical support of the OpenCV~3.0 library. We tested the program with the Fourier transform libraries CUFFT (version provided in CUDA~v.10.2) for GPU execution and FFTW~3.3.3~\cite{FFTW05} for the CPU implementations. The code was compiled with MS compiler v120 and the compiler of CUDA~v.10.2 in Visual Studio 2019 and executed on a machine with the following specifications: 
\begin{itemize}
    \item CPU: Intel® Core™ i9-9880H
    \item 32 GB DDR4 RAM
    \item Graphic card: NVIDIA Quadro RTX 4000 with 8GB of dedicated G-RAM memory
    \item 512 GB SSD drive - PCIe, performance class 40
\end{itemize}

\begin{figure}
\includegraphics[width=1\columnwidth]{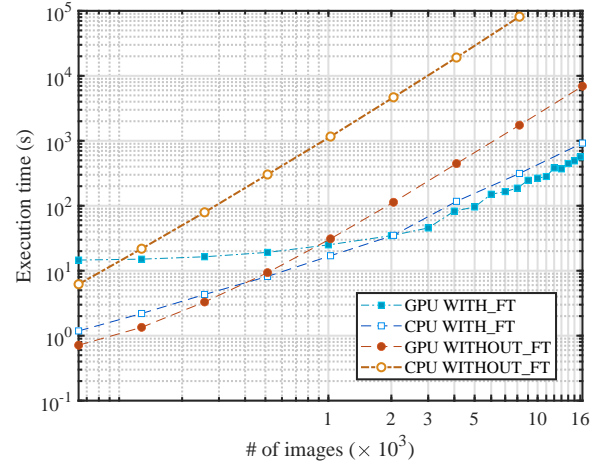}
\caption{Execution time as a function of the total number of images for images of $512\times 512$ pixels. The curves corresponding to the WITH\_FT algorithm have square markers and the curves of the WITHOUT\_FT algorithm have circular markers. The markers have colored filling for the GPU modes and have white filling for the CPU modes.}
\label{fig:execution_time_512}
\end{figure}

In our tests, we considered image sets composed by maximum $2^{14}=16384$ images with 16~bit depth and we limited the execution times to less than $10^5$~s. In the first test, we ran all the algorithms on CPU and GPU with images composed by $512\times 512$ pixels. For comparison, we made use of 8~GB of RAM for executing the program on CPU so that the CPU and the GPU could access the same amount of RAM and G-RAM, respectively. The execution times of the programs in these conditions are presented in fig.~\ref{fig:execution_time_512}, in which the times for all the four execution modes are plotted as a function of the number of images used for the test. As expected from the results reported in Ref.~\cite{Cerchiari2012}, the WITHOUT\_FT algorithm executes more than 30 times faster on GPU than CPU. The GPU hardware is also faster than the CPU in executing the WITH\_FT algorithm, but the speed-up factor never exceeds a factor of 2. Comparing the WITH\_FT with the WITHOUT\_FT scheme, the WITH\_FT scheme is faster than the WITHOUT\_FT method in processing more than $\sim1000$ images. After this threshold, both CPU and GPU versions of the WITH\_FT algorithm execute quicker than the GPU--WITHOUT\_FT implementation, reaching a maximum speed-up factor of $10-12$ for $16384$ images.
\begin{figure}
\includegraphics[width=1\columnwidth]{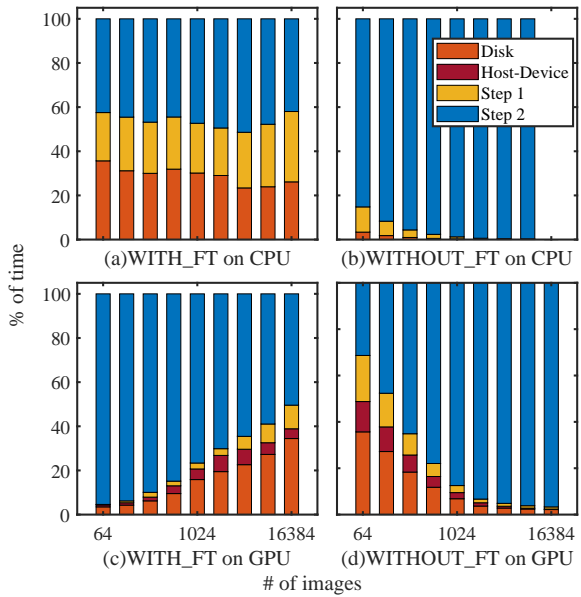}
\caption{Fractional execution time of the different tasks of the program as a function of the total number of images. The length of the colored bars represents the fractional time spent by the program to execute the different operations: \textit{disk} IO, \textit{host--device} data transfers, \textit{step 1} and \textit{step 2}. In this test, we used images of $512\times 512$ pixels. The first row (graphs (a) and (b)) presents the fractional times spent in CPU mode and the second row (graphs (c) and (d)) in GPU mode. The first column shows the fractional times of the WITH\_FT algorithm (graphs (a) and (c)) and the second column of the WITHOUT\_FT algorithm (graphs (b) and (d)). Data of the CPU WITHOUT\_FT version for 16384 images are not reported because the total execution time was exceeding $>10^5$~s.}
\label{fig:break_down_512}
\end{figure}
Figure~\ref{fig:break_down_512} presents the fractional time spent by the program in the four modes to compute the image FFT (\textit{step one}), process the time sequences (\textit{step two}) and perform memory IO operations (\textit{disk} and \textit{host-device}). The IO operations named \textit{host-device} include the data transfers between the RAM and the G-RAM and they only exists in the GPU implementations. In the figure, we normalized the fractional times by the total execution time to highlight the different workloads for executing each part of the program. As a function of increasing number of images, the workload of step two compared to the other operations remains balanced in the CPU-WITH\_FT implementation and it reduces in GPU-WITH\_FT implementation. Conversely, the WITHOUT\_FT algorithm spends more fractional time during the second step as the number of images increases both in the CPU and the GPU modes. Combining the information of figs.~\ref{fig:execution_time_512} and \ref{fig:break_down_512}, we see the advantage of the new implementation applied to the problem of calculating the structure function: the WITH\_FT algorithm is faster than the WITHOUT\_FT scheme for a large number of images as a consequence of the reduction in computational complexity in processing the time sequences of the wave vectors.

\begin{figure}
\includegraphics[width=1\columnwidth]{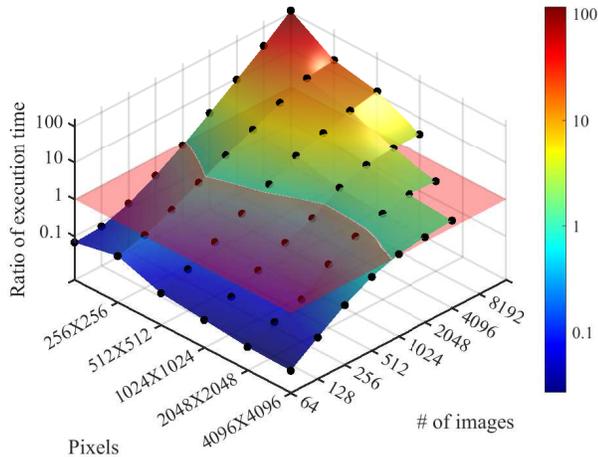}
\caption{Ratio of execution times on GPU of the WITHOUT\_FT against the WITH\_FT algorithm as a function of different number and size of images. The transparent red plane marks the condition in which both algorithm process the images in the same time.}
\label{fig:exetime_ratio_GPU_GPU}
\end{figure}

In a second test, we analyzed the execution performance of the GPU--WITH\_FT and GPU--WITHOUT\_FT algorithms for squared images of different sizes. Figure~\ref{fig:exetime_ratio_GPU_GPU} presents the ratio of execution times between GPU--WITH\_FT over GPU--WITHOUT\_FT for different number of images and image size. In analogy to the $512\times 512$ example, the WITH\_FT method is faster than the WITHOUT\_FT technique for more than $\sim500-1000$ images. The red plane in the figure marks the condition in which both algorithms complete execution in the same amount of time. We notice that small image sizes obtain a larger speedup gain as compared to large images. For example, images composed of $128\times128$ pixels obtain up to a $\sim100$ speed-up gain in the execution time, against only $\sim4$ of the $1024\times1024$. In fact, the number of pixels per image affects the load of data-transfer and FFT of the images (step one and memory operations). While processing large images, both the WITH\_FT and WITHOUT\_FT algorithm must spend an increasingly large fraction of time to prepare the time sequences before their analysis. Considering for example the WITH\_FT at processing 16384 images, the first step and memory IO occupy 44\% of the execution time with images composed of $1024\times1024$ pixels, and they occupy 62\% of the execution time for the $2048 \times 2048$ pixel pictures. Two reasons determine this fractional increase of the time spent by the program to compute the first step and performing memory IO operations. First, calculating the bidimensional FT requires more time for larger image size. Second, as mentioned, the FFTs are calculated several times if the wave vectors components of all the images exceed the available memory (see Appendix C for further details). The latter effect can be reduced by adopting larger memory areas to store the image FFTs. For this reason, as a final test, we executed the CPU--WITH\_FT algorithm with images of $512\times512$ pixels releasing to the program 23 GB of RAM. Compared to the previous tests in which the RAM was limited to 8~GB, we obtained a speedup factor of 2 thanks to the larger available memory area. In fact, the image's FT are recalculated six times by using 8~GB of RAM but only two times by using 23~GB of RAM.

\section{Conclusion}\label{sec:conclusion}

In this work, we presented a new algorithm to process image sets of Differential Dynamic Microscopy experiments which is based on the temporal FFT of the images. We developed a program to compare our new approach with the state-of-the-art algorithm reported in Ref.~\cite{Cerchiari2012}. The program features the possibility to execute both algorithms on GPU and CPU hardware. While the old approach executes $\sim30$~times faster in GPU mode compared to CPU mode, the new method performs similarly on GPU and CPU. Comparing the two algorithms, the new method outperforms the GPU-based implementation of the old scheme at processing $\gtrsim1000$ images. We recorded a 10-fold reduction in the execution time of the program by processing images of $512 \times 512$ pixels with the new algorithm in place of the old one. This speedup is obtained by both the CPU and the GPU implementations of the new scheme against the GPU-base implementation of the old algorithm. Thus, a computer with no GPU hardware can process the same images $\sim300$ times faster by applying the new algorithm instead of the old one.

The program makes use of the RAM memory to avoid recalculating the bi-dimensional FFTs of the images. Therefore, if the image FTs cannot be saved completely on the RAM, the execution time of the program increases. To reduce the impact on the execution time of large data-sets, in the final program, we implemented the possibility to reduce the number of wave vectors that are considered in calculating the structure function $d\left(m\right)$. The user may decide to limit the analysis to low-frequency wave vectors; thus, reducing the effective size of the images' FFTs to be stored on the local memory. This feature can help reducing the computation time of the program in DDM applications in which the high-frequency spatial components of the images can be neglected. \\

\textit{Acknowledgements.} This work has received funding from the European Union’s Horizon 2020 research and innovation program under the Marie Skłodowska-Curie grant agreement No 801110 and the Austrian Federal Ministry of Education, Science and Research (BMBWF).
It reflects only the author's view, the EU Agency is not responsible for any use that may be made of the information it contains. We thank Julia Braun for proofreading the manuscript.

\section{Authors contributions}
G.C. designed the algorithm. N.M. and G.C. developed the program. N.M. acquired and analyzed the data. All the authors were involved in the preparation of the manuscript. All the authors have read and approved the final manuscript.

\bibliography{bibliography}

\section{Appendix A. Number of threads}
\renewcommand{\thefigure}{A.\arabic{figure}}
\setcounter{figure}{0}
\renewcommand{\theequation}{A.\arabic{equation}}
\setcounter{equation}{0}
As discussed in the main text, the WITHOUT\_FT algorithm performs efficiently by adopting a parallel computing scheme on GPU hardware. This does not apply to the WITH\_FT scheme. To analyze the influence of parallel computing on the execution time of the WITH\_FT algorithm, we implemented the WITH\_FT method with a user-configurable number of threads both in the CPU mode and the GPU mode. The number of threads in the CPU mode refers to the number of threads spawned to execute a particular task, such as the FFT operations. In the GPU mode, the number of threads selects the amount of CUDA threads of each CUDA kernel. In both CPU and GPU modes, the number of threads also determines the number of time sequences that are processed in parallel. fig.~\ref{fig:ratio_tim_512} presents the total execution times of the program as a function of the different number of threads for 8192 and 16384 images of $512 \times 512$ pixels. Parallel computing achieves a minimal or detrimental impact on the speed-up factor in the CPU mode. In GPU mode, the performance gain saturates at around 32 threads for the CUDA kernels with a peak performance at 256 GPU--threads. Based on the results of this test, we selected the optimal number of two CPU-threads and 256 GPU-threads for all other tests presented in this work. 

\begin{figure}
\includegraphics[width=1\columnwidth]{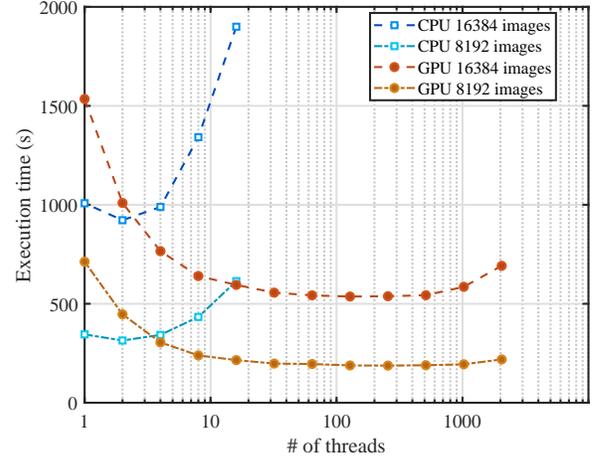}
\caption{Execution time of the WITH\_FT algorithm as a function of different number of threads. The total number of images is indicated in the legend together with the hardware platform of execution. The images size is $512 \times 512$ pixels.}
\label{fig:ratio_tim_512}
\end{figure}

\section{Appendix B. Analysis of the time sequences}
\renewcommand{\thefigure}{B.\arabic{figure}}
\setcounter{figure}{0}
\renewcommand{\theequation}{B.\arabic{equation}}
\setcounter{equation}{0}
In the main text we presented a new algorithm for the analysis of DDM images. This method approaches the analysis of DDM images via eq.~\ref{eq:newalg} and by using the fast Fourier transform (FFT) algorithm both in 2D-space and, eventually, in time. In the second step, we applied eq.~\ref{eq:newalg} to each time sequence of the wave vectors separately. In this appendix, we describe how we implemented the computation on a single time sequence by using eq.~\ref{eq:newalg} to obtain the final result.

We split the calculation on the time sequence in two parts $d\left(m\right)=d_a\left(m\right)+d_c\left(m\right)$ where
\begin{align}
    \label{eq:newalg_da}
    d_a\left(m\right)&=\frac{1}{N-m}\sum_{n=m}^{N-1}\left(\lvert \tilde{I}_{n-m}\rvert^2+\lvert \tilde{I}_{m}\rvert^2\right) \; , \\
    \label{eq:newalg_dc}
    d_c\left(m\right)&=\frac{2}{N-m}\sum_{n=m}^{N-1}\textrm{Re}\left( \tilde{I^*}_{n-m}\tilde{I}_{m}\right) \; .
\end{align}

The term $d_a\left(m\right)$ is calculated by using the following iterative formula:
\begin{equation}
    d_a\left(N-n-1\right)=\frac{n}{n+1}d_a\left(N-n\right)+\frac{\lvert I_{n}\rvert^2+\lvert I_{N-n-1}\rvert^2}{n+1} \;,
\end{equation}
where the index $n\in\left[0,N-1\right]$.

The term $d_c\left(m\right)$ expresses the autocorrelation of the time sequences. We implemented the autocorrelation by using the FFT in time considering two requirements. First, the maximum performance gain is expected if the support points of the time sequences are a power of two to take advantage of the FFT speedup. Second, the summation over $n$ of eq.~\ref{eq:newalg_dc} takes into account only $N-m$ pairs of $\tilde{I}_{n}$ functions which is incompatible with evaluating the FFT directly on $N$ support points. The incompatibility emerges because of the periodical boundary conditions imposed by the FFT. To meet both requirements, we zero-padded the time sequences to $N_2$ support points solving the equation:
\begin{equation}
    \log_2{N_2} = \lceil\log_2{\left(N\right)}\rceil + 1 \; ,
\end{equation}
where ``$\lceil\ldots\rceil$'' denotes the ceiling operation. This padding operation allows us to use the FFT and calculate exactly eq.~\ref{eq:newalg} without any influence caused by the periodical boundary conditions. The calculation of $d_c\left(m\right)$ for a single time sequence can be broken down in the following operations.
\begin{itemize}
    \item The time sequence is zero-padded to $N_2$ complex support points.
    \item The padded sequence is Fourier transformed by FFT in time.
    \item Each element of the FFT is squared in modulus obtaining the power spectrum of the time sequence.
    \item The inverse FFT is applied to the power spectrum.
    \item The real part of the first N elements of the resulting time sequence are normalized by the ramp vector $1/\left(N-m\right)$.
\end{itemize}
Finally, $d_a\left(m\right)$ and $d_c\left(m\right)$ are added together to obtain $d\left(m\right)$.

\section{Appendix C. Group execution}
\renewcommand{\thefigure}{C.\arabic{figure}}
\setcounter{figure}{0}
\renewcommand{\theequation}{C.\arabic{equation}}
\setcounter{equation}{0}
\begin{figure}
\includegraphics[width=1\columnwidth]{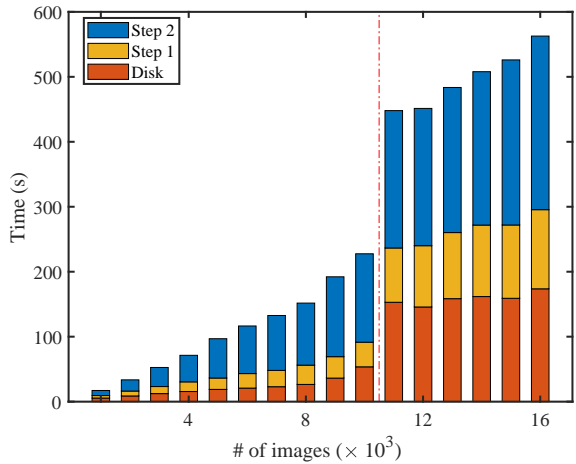}
\caption{Execution time of the WITH\_FT algorithm on CPU hardware on images of $512 \times 512$ pixels and 23~GB of RAM. The vertical line marks the crossing point in which the algorithm divides the execution from one into two groups.}
\label{fig:time_break_cpu}
\end{figure}
The program described in this work splits the calculations in groups if the data of all the wave vector components for all the images exceeds the available storage memory. The method WITHOUT\_FT uses a first-in-first-out (FIFO) memory scheme already described in Ref.~\cite{Cerchiari2012}. This approach aims to calculate groups of complete $d\left(m\right)$ matrices. The WITH\_FT algorithm, instead, operates sequentially on different groups of wave vectors for all the $d\left(m\right)$ and saves the partial results of each group on the hard-drive. The partial results are merged at the final stage of the program. In practice, in both algorithms, the images are loaded and Fourier transformed one time for each group because only a part of the FFTs data can be saved on the local memory. The impact of repeating these operations over the entire execution time is presented in fig.~\ref{fig:time_break_cpu}, in which we present the execution time as a function of the number of images to process. In this test, we executed the WITH\_FT algorithm on CPU hardware with images of $512 \times 512$ pixels by releasing to the program 23~GB of RAM. In the figure, the vertical red line marks the crossing point from one-group to two-group execution. We see that the time spent by the program in memory operations and step one suddenly doubles by crossing the line because the bi-dimensional FFTs of the images and the corresponding IO operations must be executed two times instead of only one.

\end{document}